\documentclass[runningheads]{llncs}
\usepackage{graphicx}

\usepackage{listings}
\usepackage{amsmath}
\usepackage{amsfonts}
\usepackage{algorithm}
\usepackage{algorithmic}
\usepackage{multirow}
\usepackage{colortbl}
\usepackage{color}
\usepackage{makecell}
\usepackage{longtable}
\usepackage{pdfpages}
\usepackage{booktabs}
\usepackage{pdflscape}
\usepackage[acronym,toc]{glossaries}
\usepackage{setspace}
\usepackage{tabularx}
\usepackage{float}
\usepackage{subfigure}
\begin{document}
\title{Interpretable and High-Performance Hate and Offensive Speech Detection}
\author{Marzieh Babaeianjelodar\inst{1} \and
Gurram Poorna Prudhvi \and
Stephen Lorenz\inst{2} \and Keyu Chen \inst{1} \and Sumona Mondal \inst{2} \and Soumyabrata Dey \inst{2} \and Navin Kumar \inst{1}}
\authorrunning{Babaeianjelodar et al.}
\institute{Yale University, New Haven, CT 06520, USA
\email{\{marzieh.babaeianjelodar,keyu.chen,navin.kumar@yale.edu\}}\\ \and
Clarkson University, Potsdam, NY 13699, USA\\
\email{\{lorenzsj,smondal,sdey\}@clarkson.edu}}
\maketitle              % typeset the header of the contribution
\begin{abstract}
The spread of information through social media platforms can create environments possibly hostile to vulnerable communities and silence certain groups in society. To mitigate such instances, several models have been developed to detect hate and offensive speech. Since detecting hate and offensive speech in social media platforms could incorrectly exclude individuals from social media platforms, which can reduce trust, there is a need to create explainable and interpretable models. Thus, we build an explainable and interpretable high performance model based on the XGBoost algorithm, trained on Twitter data. For unbalanced Twitter data, XGboost outperformed the LSTM, AutoGluon, and ULMFiT models on hate speech detection with an F1 score of 0.75 compared to 0.38 and 0.37, and 0.38 respectively. When we down-sampled the data to three separate classes of approximately 5000 tweets, XGBoost performed better than LSTM, AutoGluon, and ULMFiT; with F1 scores for hate speech detection of 0.79 vs 0.69, 0.77, and 0.66 respectively. XGBoost also performed better than LSTM, AutoGluon, and ULMFiT in the down-sampled version for offensive speech detection with F1 score of 0.83 vs 0.88, 0.82, and 0.79 respectively. We use Shapley Additive Explanations
(SHAP) on our XGBoost models' outputs to makes it explainable and interpretable compared to LSTM, AutoGluon and ULMFiT that are black-box models. 
\end{abstract}
\section{Introduction}
\label{sec:gv1}
Millions of people around the globe use social media such as Facebook, Twitter, and Youtube as news sources because of its easy access and low cost \cite{monti2019fake}. With the widespread reach of social media platforms, people can easily share their thoughts and feelings to a large audience. However, social media platforms can be a double-edged sword. While social media can assist in making peoples lives better and providing an environment for people to talk freely, some posts by individuals can contain hateful or offensive speech. Hate speech is language used to express hatred towards a targeted individual or group \cite{davidson2017automated}, while offensive speech is strongly impolite, rude or vulgar language expressed towards an individual or group \cite{davidson2017automated}. Hate and offensive speech can easily proliferate on social media and is often not easy to characterize. Hate and offensive speech can impact individuals' self-esteem and mental health, create a hostile online environment and in extreme cases, incite violence. Hate and offensive speech can also marginalize vulnerable communities further worsening outcomes for such groups. Thus, researchers have worked on detection of hate and offensive speech \cite{davidson2017automated,founta2018large}.

Different works around text classification have been done such as spam detection \cite{jindal2007review}, financial fraud detection \cite{ngai2011application}, and emergency response \cite{caragea2011classifying} using Natural Language processing (NLP) techniques. In our work we would like to draw attention to a certain type of text classification task in social media. Machine learning systems which have been used to detect hate and offensive speech in a range of cases. For example, Facebook makes decisions to remove particular posts related to hate speech that are targeted at the black community with the help of machine learning systems trained on user based flagged content \cite{dwoskin_tiku_kelly_2020}. In an article written by \cite{rosenberg_2021} in 2018, Washington Post then reported that facebook falsely flagged Declaration of Independence as hate speech which shows how challenging the detection of hate speech can be for the biggest social media platforms and shows the impact of these algorithms on our daily lives. In this regard, different models have been built for hate and offensive speech detection but the majority of these are not interpretable or explainable. Interpretability is the ability to determine cause and effect from a machine learning model. Explainability is the knowledge of what a node represents and its importance to a model's performance. Interpretability and explainability can assist social media platforms in providing justification for banning an individual for hate speech, helping platform users to understand and mitigate their own inappropriate behavior. The lack of explainability and interpretability in such systems may also lead to user distrust or movement to alternative, unregulated platforms where hate speech proliferates. For example, alternative platforms like Parler were used to plan and promote terrorist attacks \cite{hannah_allam_2021}. Such views are echoed by FAT-ML's principles for accountable algorithms where algorithmic decisions and the related data driving those decisions should be explained to users and other stakeholders in non-technical terms. Similarly, hate detection models may inherit the biases of their data. For example, if training data is politically biased, users who make controversial but non-offensive posts may get banned from social media platforms. 

While there has been significant work around hate speech detection \cite{mathew2020hatexplain,caselli2020hatebert} limited work has delineated the differences between hateful and offensive speech. Because of the overlapping definitions of hate and offensive speech it is often hard to distinguish between them. In the social media environment, this distinction is key. For example, platforms may want to issue stronger cautions for hate speech compared to offensive speech \cite{davidson2017automated}. Thus, automated hate detection systems for social media should not only be highly accurate but also explainable and interpretable, and able to distinguish between hate and offensive speech. Several machine learning approaches, especially neural network systems, used in detecting hate and offensive speech demonstrate high performance but are not often explainable or interpretable \cite{bunde2021ai}. Such hate and offensive speech detection models do not allow us to see which features contribute to the detection of a hate or offensive speech, key to building trust in such systems. We use a decision-tree based ensemble algorithm which uses the gradient boosting framework called XGBoost. It has shown better results across many Kaggle competitions, real world applications and at times gave better results than neural network based approaches \cite{reinstein_2017}. Since the structure of the algorithm is based on decision trees, we can make use of the predictive power of the features utilized for the classification and infer the predictions based on the features. Thus, when using XGBoost over neural networks, a hate and offensive speech detection model is more explainable and interpretable in describing which features have led to a particular class detection. We build a hate and offensive speech detection model to differentiate between hate and offensive speech using Twitter data. Our work differs from other studies that have tried to differentiate between hate and offensive speech by creating a more accurate, explainable, and interpretable hate and offensive speech detection system \cite{davidson2017automated}. To construct our model, we have used an optimal set of feature sets combined with our own added features \cite{davidson2017automated}. Past research has used a limited number of features and we extend the literature by considering a comprehensive set of features to improve model performance, explainability, and interpretability, and ability to distinguish between hateful and offensive speech. We propose the two research questions (RQ). RQ1: How can we make hate and offensive speech detection models more explainable and interpretable? and RQ2: How can we improve the performance of hate and offensive speech detection models?

\section{Background and Related Work}
\subsubsection{Hate and Offensive Speech Detection Models}
There exist several hate and offensive speech detection models \cite{mathew2020hatexplain}. Most of these models are based on supervised machine learning systems such as support vector machines, random forest, logistic regression and decision trees \cite{fortuna2018survey}. \cite{davidson2017automated} predict hateful, offensive, and neither speech from crowdsourced labels. However, developing appropriate features to detect hateful and offensive speech is often complex. \cite{davidson2017automated} use TF-IDF, part of speech (POS) tagging, readability scores, sentiment, binary and count indicators for hashtags, mentions, retweets, URLs, and a number of characters, words, and syllables in each tweet as features for the classifier. Neural networks have been used more recently for hate speech detection \cite{DBLP:journals/corr/abs-1802-05365} because of their high performance. For example, \cite{cao2021deephate} developed a deep hate model which combines different models and features to detect hate speech. They feed feature embeddings into neural network models allowing the models to learn semantics, sentiment, and topics. Their model detects hate speech more accurately compared to other models. DeepHate has shown to outperform eight other models on four tasks in all but one case \cite{cao2021deephate}. 

\subsubsection{Model Interpretability and Explainability}
Machine learning models are being implemented in an increasing range of areas, such as health, criminal justice systems \cite{berk2021fairness}, and social media platforms. However, the explainability and interpretability of these complex models is not often reported \cite{gilpin2018explaining}. Limited explainability and interpretability reduces trust in machine learning models and may increase bias \cite{zachary2016mythos}. For example, neural networks such as long short term memory (LSTM) have impressive performance, but they are often not explainable \cite{linardatos2021explainable}. For example, such models may misdiagnose cancer in a patient or incorrectly predict recidivism risk \cite{chouldechova2017fair}. To reduce biases in machine learning models, deploying explainable and interpretable models is paramount. Such models can aid social media companies in explaining their decisions to users, building trust and allowing users to correct their future behavior. We provide some examples of improving model explainability and interpretability. \cite{zaidan2007using} introduced the use of human annotators to highlight parts of the text to support their model's labeling decisions. \cite{yessenalina2010automatically} developed an automatic rationale generator which can replace the human annotation method. In the process of developing more interpretable models \cite{mathew2020hatexplain} developed the first benchmark hate speech dataset containing human level descriptions. \cite{mathew2020hatexplain} show that existing models may have high performance but do not usually have high explainability. Their findings demonstrate that models with human rationales in the training process tend to be less biased. 

\subsubsection{Models of Interest}
XGBoost is a scalable implementation of a gradient boosting framework by \cite{friedman2001greedy}, which decreases model errors. XGBoost converts weak learners to strong learners in a sequential way, leading to each model correcting the previous model. Because of the different capabilities of XGBoost, this model is a good fit for machine learning systems. In particular, XGBoost exhibits better performance compared to deep learning models because of missing values handling, and in-built cross-validation features \cite{stephens-davidowitz_2012,mathew2020hatexplain}. In addition, XGBoost performs very well on huge datasets in a shorter period of time compared to neural networks because of its parallel processing capability. Because the problem statement requires interpretability for detecting hate and offensive speech, XGBoost is an apt choice since it shows the weights given to different features as opposed to neural networks which are limited on interpretability. Lastly, XGBoost also performs well on imbalanced data.

Long Term Short Memory (LSTM) \cite{hochreiter1997long} is a type of recurrent neural network (RNN) capable of learning long-term memory and short-term memory. LSTM was introduced since RNN is not capable of learning short-term memory, and is widely used in language generation, voice recognition, and optical character recognition (OCR) models. Forward neural networks (FNN) are useful for simple predictions such as predicting if an image is a dog or a cat, however, RNNs are good for predicting words in a sentence since RNNs capture the sequence of inputs. In comparison with RNN, LSTM can capture the longer sequence, making LSTM ideal for hate/offensive speech classification on tweets. 

AutoGluon is an automatic machine learning library, introduced by Amazon AWS \cite{erickson2020autogluon} which automates machine learning and deep learning methods that train images, texts and tabular datasets. AutoGluon automatically cleans the dataset, trains and predicts with less developer-level involvement. We chose the AutoGluon library since it is easy to use and it has good performance on supervised machine learning models. AutoGluon uses ensemble modeling and stacking the models in different layers. Thus, it is a good benchmark and prototype for comparing other models such as XGBoost and LSTM. 

Universal Language Model Fine Tuning (ULMFiT) \cite{howard2018universal} is a generic transfer learning technique which is applied to NLP tasks. This model has been implemented in the fastai deep learning library and has shown good performance on various nlp tasks such as text classification \cite{merity2017regularizing}. The language model is initially trained using Wikitext-103 which consists of 28595 Wikipedia articles and 103 million words. This step is called the General Domain Language Model (LM) pre-training. Then in the next step the model will be further fine tuned for the text classification which is called the Target Task LM Fine Tuning. Finally, a target task classifier is built based on the previous steps. 

\section{Data}
We used the dataset from \cite{founta2018large}, due to its size and reliability. The data is comprised of 80,000 annotated tweets, labeled as hate speech, offensive speech, or neither. The data distribution of \cite{founta2018large} is not balanced as the tweets were generated by real users. In general, we find less hate and offensive speech compared to neither on twitter. The original distribution of the tweets were: Neither: 53731, Offensive: 27229, Hate: 5006. We started with different data analysis techniques to understand and analyze the data which are explained in the following.

\subsubsection{Word clouds}
\begin{figure}[h]
    \centering
    \subfigure[Hate Speech\label{fig:wcloud0}]{{\includegraphics[height=2cm, width=3.8cm]{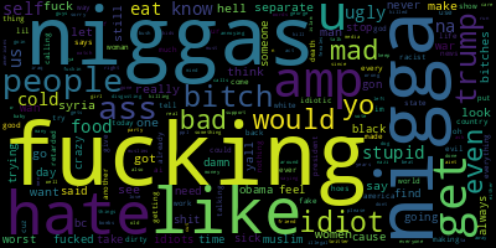}}}\qquad
    \hspace*{-0.5cm}
    \subfigure[Offensive Speech\label{fig:wcloud1}]{{\includegraphics[height=2cm, width=3.8cm]{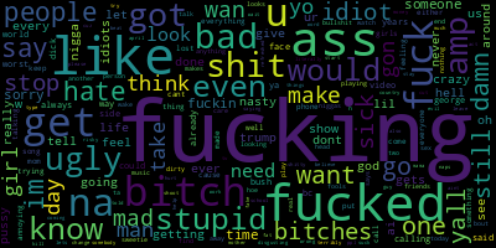}}}\qquad
    \hspace*{-0.5cm}
    \subfigure[Neither\label{fig:wcloud2}]{{\includegraphics[height=2cm, width=3.8cm]{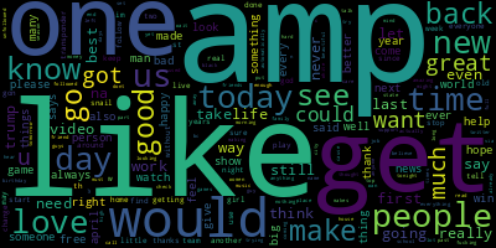}}}%
    \caption{Word clouds for various forms of speech.} \label{fig:clouds}
\end{figure}

The word clouds in figure \ref{fig:clouds} shows the words mostly used in hate, offensive and neither classes of the dataset. The hate speech word cloud \ref{fig:wcloud0}, shows the frequent usage of the hatred words such as ``niggas'', ``fucking'', ``bitch'' and the offensive speech word cloud \ref{fig:wcloud1} shows the frequent usage of offensive words such as ``fucking'', ``stupid'', ``shit'' which can be overlapping sometimes but can be different in many cases which makes building a hate and offensive detection model challenging. Finally, the neither class \ref{fig:wcloud2} word cloud shows neutral words such as ``like'', and ``get''.

\subsubsection{Keyword extraction}
We used another method to analyze the words in each class since word cloud did not completely distinguish the hate and offensive class. We used a method called KeyBERT \cite{grootendorst2020keybert}, a keyword extraction technique based on DistilBERT which is a smaller and lighter version of the BERT model. We performed keyBERT on class hate and class offensive. The top keywords associated with class hate were ``nigga'' and ``nigger'' while the top keywords associated with class offensive were ``bitch'' and ``douchebag'' which shows the line between these two classes better.

\section{Method}
\subsubsection{Feature Selection} 
In creating the XGBoost model we started with a base set of features as paper \cite{davidson2017automated} to construct our model. For each new feature addition, we analyze the impact of that particular feature on the performance of the model. After testing different features and analyzing the performance of the model, we derived with an optimal set of features which produced the best model performance as shown in Table ~\ref{table:Features}. The features in the table are described below.

\subsubsection{Model performance}
To analyze the performance of our model in each step of our feature selection, we use F1 score instead of accuracy. F1 score is a weighted average of precision and recall which prevents the majority class bias. We have performed cross validation using sklearn cross\_val\_score. For our use case we have used K-FOLD cross validation with k=5 and verified the results across the folds and reported the average F1 score. 

\subsubsection{Sentiment} 
When discussing detecting hatefulness of a certain sentence, one of the main features that can be added to a model is finding the polarity of that piece of text by sentiment analysis. Therefore, the first feature added in Table ~\ref{table:Features} is sentiment score, used in similar work \cite{davidson2017automated}. Tweets that contain hate speech tend to have higher negative sentiment compared to other tweets, thus sentiment is an appropriate feature to detect the hate class. We use the VADER \cite{hutto2014vader} Sentiment Analyzer to assign sentiment score. With the addition of sentiment, the F1 scores for offensive speech, hate speech, and neither are 0\%, 62\%, and 84\% respectively.  

\subsubsection{Part of Speech Tagging and Named Entity Recognition} 
We added part of speech (POS) tagging using the SpaCy library. The SpaCy library also has a fast Named Entity Recognition (NER) system that we used. The SpaCy library was used to label named entities produced from the text such as a person, a country, a place, an organization, etc. After adding these features to the model, the model shows an improved F1 score of 86\% for neither class, 70\% for the hate class, and 17\% for the offensive class. 

\subsubsection{Hashtags and Mentions} 
We included the number of hashtags and mentions in the tweets by using the python preprocessing library called preprocessor. When we added those features to the model, neither class showed an F1 score of 87\%, with similar improvements for hate class (73\%) and offensive class (18\%). 

\subsubsection{Emojis and Other Symbols}  
Social media platforms have seen an increasing usage of emotion symbols such as emojis \cite{kejriwal2021empirical} \footnote{http://www.iemoji.com/} especially in twitter. These emojis can carry positive or negative emotion contents such as joy, celebration, anger, disgust and etc. We added these text symbols (emojis) and other text symbols such as, exclamation points, question marks, words in all caps, and periods as features. To extract these symbols we used the regex library in Python. After adding these features to the model, the offensive class F1 score was 88\%, the hate class was 75\%, and neither class was 22\%.

\subsubsection{TF-IDF Analysis} 
We calculated the TF-IDF metric, which is a statistical measure that stands for term frequency-inverse document frequency. We used the scikit-learn package for calculating TF-IDF. 
TF-IDF is calculated by multiplying two metrics, the frequency of a word inside a document (TF) and the inverse document frequency of the word in the whole documents (IDF). We calculated the TF-IDF for the POS tags. After adding these sets of features to the model, the offensive class F1 score was 75\%, the hate class was 88\%, and neither class was 97\%.

\subsubsection{XGBoost Model Building}
For building the XGBoost model, we pre-processed the data and read the dataset with the features included. We split the dataset into 20\% and 80\% test and train subsets. Because of the unbalanced nature of our dataset, we used a class weight parameter in the model. We then performed hyper parameter tuning using the validation dataset. We tested the model and obtained the optimal hyper parameters that led to higher F1 scores for the three classes.

\subsubsection{LSTM}
To compare our XGBoost model with neural networks we develop an LSTM model with the \cite{founta2018large} dataset. Using the pre-trained model we create an embedding matrix that contains word vectors for all the unique words in the cleaned tweets dataset. Using this embedding matrix, and the Keras embedding layer, we convert the input tweets to embeddings and pass them to the neural network. We split the data into 20\% and 80\% test and train subsets respectively. All the tokens were transformed into word vectors using the Google News pre-trained corpus. The word vector representations were then passed into the LSTM neural network. We used Keras to create the structure of our model. Our model consists of an embedding layer, conv1d layer, followed by MaxPooling1D, batch normalization, spatial dropout1d, bidirectional LSTM, and an output layer.

\subsubsection{AutoGluon}
We use the AutoGluon model to serve as a benchmark for comparing our XGBoost, LSTM, and ULMFiT models. We split the dataset into 20\% and 80\% test and train subsets. We set the hyper parameters to use the multi-model feature, where it trains multiple models such as the text predictor model, and combines them through weighted ensemble or stack ensemble methods depending on its performance. Finally we allow AutoGluon to select the best performance algorithm based on test and validation scores.

\subsubsection{ULMFiT}
We first clean the \cite{founta2018large} dataset, and split the dataset into 20\% and 80\% test and train subsets. We then use the text data loader object to adjust the format of the input file for the model. We create our ULMFiT model using the Fastai deep learning library. We use the text classifier learner from the Fastai library which uses the Averaged Stochastic Gradient Descent (ASGD) Weight-Dropped LSTM \cite{merity2017regularizing} model. We use a pre-existing pretrained model (trained on Wikitext 103) and perform fine tuning with our dataset on it. Finally, we tuned the model to get the best performance and our best performance classifier was built based on 20 epochs, and 0.003 learning rate.      

\begin{table}
  \caption{F1 Score of the three classes with the inclusion of different feature sets with XGBoost. To readability, we used the following abbreviations in the figure below: Sent = Sentiment, Hash = Hashtag, Men = mention, Symb = Symbol.  } \label{table:Features}
  \begin{tabular}{ccccc}
    \toprule
    Row & Features Used & Neither & Offensive & Hate\\
    \midrule
      (1) & Sent & 0.84 & 0 & 0.62 \\
      (2) & Sent, POS + NER & 0.86 & 0.17 & 0.70\\
      (3) & Sent, POS + NER, Hash + Men & 0.87 & 0.18 & 0.73\\
      (4) & Sent, POS + NER, Hash + Men, Text Symb & 0.88 & 0.22 & 0.75 \\
      (5) & Features used in Row (4), POS + TF-IDF & 0.97 & 0.75 & 0.87\\
    \bottomrule
    \end{tabular}
\end{table}

\subsection{SHAP-based Explanations of the model}
To understand the main contributions behind how a model predicts or labels is key to trusting models. On the other hand the best machine learning systems are complex to understand, therefore, to use and trust such systems some methods have been introduced such as the SHAP (SHapley Additive exPlanations) which is a unified framework for interpreting predictions \cite{SHAP}. SHAP uses an importance value for each feature contributing to the prediction of the model. SHAP's approach is based on the shapley values drawn from game theory and are applied to machine learning models for explainability. In our work, we used the Shapley Additive Explanations (SHAP) force plot to interpret the tweets in our dataset. For the purpose of the transparency of our model, we use the SHAP tree explainer. We show an examples from the tweets to see how the different features chosen are impacting the decisions made by the model to classify the tweet in a certain category. 

\begin{figure}
  \includegraphics[width=\linewidth]{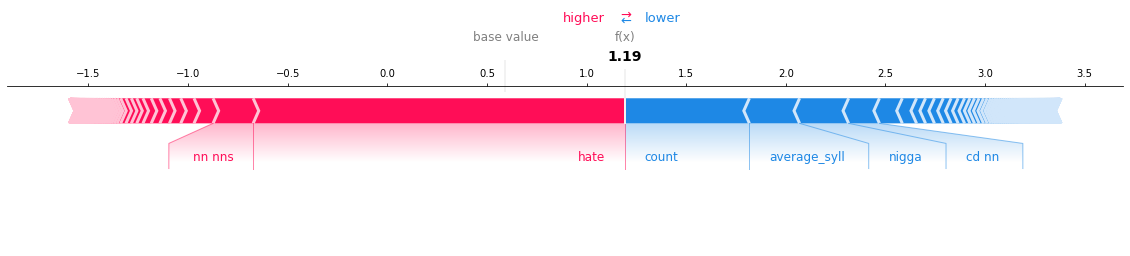}
  \caption{Shap for the tweet "if you still hate this nigga" labeled as class hate.}
  \label{fig:Shap3}
\end{figure}

In the SHAP model the tweet ``if you still hate this nigga xxx http:xxxx'' shows the degree of this tweet belonging to class 0 (hate) is 1.19 shown in \ref{fig:Shap3}. In the figure, we see that our SHAP importance value is higher than the base value which means certain features such as the word hate and POS nn (nouns) are contributing to the hate class while other features such as count and average-syl are not contributing to the tweet being categorized as the hate class. 

\section{Results}
We measure the models based on F1 score presented in Table \ref{table:Models_Analysis}. F1 score is used because our three classes are different in terms of size. In the hate class, XGBoost outperforms LSTM (F1 score: 0.75 vs 0.38), AutoGluon (F1 score: 0.75 vs 0.37) and ULMFiT (F1 score: 0.75 vs 0.38). For the neither and offensive classes, all models have similar scores. 

\begin{center}
\begin{table}[h]
    \caption{F1 Score of the three classes for XGBoost, LSTM, AutoGluon, and ULMFiT.\label{table:Models_Analysis}}
    \begin{tabularx}{\columnwidth}{*{4}{>{\centering\arraybackslash}X}}
        \toprule
        Model & Neither & Offensive & Hate  \\
        \midrule
        XGBoost & 0.97 & 0.87 & \textbf{0.75} \\
        LSTM & 0.96 & 0.91 & 0.38 \\
        AutoGluon & 0.96 & 0.90 & 0.37 \\
        ULMFiT & 0.95 & 0.89 & 0.38 \\
        \bottomrule
    \end{tabularx}
\end{table}
\end{center}

As the dataset is imbalanced, we down-sampled the classes to make them all equal sizes. We down-sampled the hate and neither classes to the minority class size of approximately 5000 tweets. We present results for the down-sampled dataset in Table ~\ref{table:downsample}. In the offensive and hate class, XGBoost performed relatively better than the other models. In the neither class, XGBoost performed better than LSTM and ULMFiT but is quite similar to AutoGluon.

\begin{table}[h]
\caption{F1 Score of the three classes for XGBoost, LSTM, AutoGluon, and ULMFiT after down sampling.\label{table:downsample}}
\begin{tabularx}{\columnwidth}{*{4}{>{\centering\arraybackslash}X}}
    \toprule
    Model & Neither & Offensive & Hate  \\
    \midrule
    XGBoost & 0.90 & 0.83 & \textbf{0.79} \\
    LSTM & 0.85 & 0.80 & 0.69 \\
    AutoGluon & 0.90 & 0.82 & 0.77 \\
    ULMFiT & 0.83 & 0.79 & 0.66 \\
    \bottomrule
\end{tabularx}
\end{table}

\section{Discussion and Conclusion}
In this work, we argue that hate and offensive language should be detected across social media platforms to lower the possibility of minority groups getting attacked based on ethnicity, religion, and disability especially in countries which advocate freedom of speech. Although people might have the opportunity to express their opinion by law in a country, they might be silenced due to the prejudice that exists towards a certain group. We discuss that while differentiating between hate and offensive speech can be challenging and could even be at times overlapping, hate and offensive language can have different meanings and can have different affects on people and should be treated differently which means social media platforms should come up with algorithms that can differentiate these two categories of languages. Categorizing offensive speakers as hate speakers can lead to falsely banning a lot of people in social media platforms. To assist in coming up with models that can detect the two types of languages, we come up with our own XGBoost based high performance model that can detect hate speech and offensive speech. We compare our XGBoost based model with other state of the art models such as LSTM, AutoGluon and ULMFiT. We show that for the entire mentioned dataset, for detecting hate speech, XGBoost outperformed LSTM, AutoGluon, and ULMFiT. Upon down sampling our data, in the hate and offensive classes, XGBoost performed better than the other models. In addition to superior performance, unlike LSTM, AutoGluon, and ULMFiT, XGBoost also offers explainability and interpretability in Machine Learning based platforms. More specifically, XGBoost shows the features that have been contributing for detecting the classes while neural network and language based models do not have that capability and are often black boxes. Unfortunately, with the increase of ML-based platforms the need of building these platforms have been surpassing the trust that these platforms provide to their users making a lot of these models black-boxes. Model explainability and interpretability can aid social media platforms in more effective moderation, helping users understand why they have been banned, allowing individuals to alter their behavior for future reference, thereby improving trust in such systems. 

\subsubsection{Limitations}
We did not account for other features in model development, such as demographic characteristics. In addition, we think that there are phrases used in social media platforms that are attacking certain groups in more sophisticated ways and can be labeled as hate speech but they are hard to detect and can only be detected within the context which we do not consider in this work. We were unable to incorporate adversarial testing.
\bibliographystyle{splncs04}
\bibliography{citations}
%Author, F.: Article title. Journal \textbf{2}(5), 99--110 (2016)

%Author, F., Author, S.: Title of a proceedings paper. In: Editor,
%F., Editor, S. (eds.) CONFERENCE 2016, LNCS, vol. 9999, pp. 1--13.
%Springer, Heidelberg (2016). \doi{10.10007/1234567890}

%Author, F., Author, S., Author, T.: Book title. 2nd edn. Publisher,

%Author, A.-B.: Contribution title. In: 9th International Proceedings
%on Proceedings, pp. 1--2. Publisher, Location (2010)

%LNCS Homepage, \url{http://www.springer.com/lncs}. Last accessed 4
%\end{thebibliography}
\end{document}